\documentclass{article} 
\usepackage{nips14submit_e,times}
\usepackage{hyperref}
\usepackage{url}
\usepackage{graphicx}
\usepackage{amsmath,amssymb}
\usepackage{color}
\usepackage{algorithm}
\usepackage{multirow}
\usepackage{enumitem}

\title{Depth Map Prediction from a Single Image \\
       using a Multi-Scale Deep Network}

\author{
David Eigen$^1$ \\
\texttt{deigen@cs.nyu.edu} \\
\And
Christian Puhrsch$^1$ \\
\texttt{cpuhrsch@nyu.edu} \\
\And
Rob Fergus$^{1,2}$ \\
\texttt{fergus@cs.nyu.edu} \\
\And
\vspace{-1.5em} \\
$^1$Dept. of Computer Science, Courant Institute, New York University \\
$^2$Facebook AI Research
}

\newcommand{\eqn}[1]{Eqn.~\ref{eqn:#1}}
\newcommand{\fig}[1]{Fig.~\ref{fig:#1}}
\newcommand{\tab}[1]{Table~\ref{tab:#1}}
\newcommand{\secc}[1]{Section~\ref{sec:#1}}

\def\etal{{\textit{et~al.~}}}
\def\ie{{\textit{i.e.~}}}

\newcommand\enum[1]{({\it #1})}

\nipsfinalcopy

\begin{document}

\maketitle

\begin{abstract}

Predicting depth is an essential component in understanding the 3D geometry of
a scene.  While for stereo images local correspondence suffices for estimation,
finding depth relations from a \emph{single image} is less straightforward,
requiring integration of both global and local information from various cues.
Moreover, the task is inherently ambiguous, with a large source of uncertainty
coming from the overall scale.  In this paper, we present a new method that
addresses this task by employing two deep network stacks: one that makes a
coarse global prediction based on the entire image, and another that refines
this prediction locally.  We also apply a scale-invariant error to help measure
depth relations rather than scale.  By leveraging the raw datasets as large
sources of training data, our method achieves state-of-the-art results on both
NYU Depth and KITTI, and matches detailed depth boundaries without the need for
superpixelation.

\end{abstract}

\vspace{-2mm}
\section{Introduction}
\vspace{-2mm}

Estimating depth is an important component of understanding geometric 
relations within a scene.  In turn, such relations help provide richer
representations of objects and their environment, often leading to improvements
in existing recognition tasks \cite{Silberman12}, as well as enabling many further
applications such as 3D modeling \cite{Saxena08,Hoiem05}, physics and support models \cite{Silberman12},  robotics \cite{hadsell2009learning,Michels05highspeed},
and potentially reasoning about occlusions.

While there is much prior work on estimating depth based on stereo images or
motion \cite{Scharstein02ataxonomy}, there has been relatively little on estimating depth
from a \emph{single} image.  Yet the monocular case often arises in practice:
Potential applications include better understandings of the many images
distributed on the web and social media outlets, real estate listings, and
shopping sites.  
These include many examples of both indoor and
outdoor scenes.

There are likely several reasons why the monocular case has not yet been
tackled to the same degree as the stereo one.  Provided accurate image
correspondences, depth can be recovered deterministically in the stereo case
\cite{Hartley2004}.  Thus, stereo depth estimation can be reduced to developing
robust image point correspondences --- which can often be found using local
appearance features.
  By contrast, estimating depth from
a single image requires the use of monocular depth cues such as line
angles and perspective, object sizes, image position, and atmospheric effects.
Furthermore, a global view of the scene may be needed to relate these effectively,
whereas local disparity is sufficient for stereo.

Moreover, the task is inherently ambiguous, and a technically ill-posed
problem:  Given an image, an infinite number of possible world scenes may have
produced it.  Of course, most of these are physically implausible for
real-world spaces, and thus the depth may still be predicted with considerable
accuracy.  At least one major ambiguity remains, though:  the global scale.
Although extreme cases (such as a normal room versus a dollhouse) do not exist
in the data, moderate variations in room and furniture sizes are present.  We
address this using a \emph{scale-invariant error} in addition
to more common scale-dependent errors.  This focuses attention on the spatial
relations within a scene rather than general scale, and is particularly apt
for applications such as 3D modeling, where the model is often rescaled during
postprocessing.

In this paper we present a new approach for estimating depth from a single
image.  We \emph{directly regress on the depth} using a neural network with two
components: one that first estimates the global structure of the scene, then a
second that refines it using local information.  The network is trained using
a loss that explicitly accounts for depth relations between pixel locations,
in addition to pointwise error.  Our system achieves state-of-the
art estimation rates on NYU Depth and KITTI, as well as improved
qualitative outputs.

\vspace{-3mm}
\section{Related Work}
\vspace{-3mm}

Directly related to our work are several approaches that estimate depth
from a single image.  Saxena \etal
\cite{Saxena05} 
predict depth from a set of image features using
linear regression and a MRF, and later extend their work
into the Make3D \cite{Saxena08} system for 3D model generation.
However, the system relies on horizontal alignment of images, and suffers
in less controlled settings.
Hoiem \etal \cite{Hoiem05} do not predict depth explicitly, but instead
categorize image regions into geometric structures (ground, sky, vertical),
which they use to compose a simple 3D model of the scene.

More recently, Ladicky \etal \cite{Ladicky14} show how to integrate
semantic object labels with monocular depth features to improve performance;
however, they rely on handcrafted features and use superpixels to segment the
image.  Karsch \etal \cite{Karsch14} use a kNN transfer mechanism based on SIFT
Flow \cite{siftflow} to estimate depths of static backgrounds from single
images, which they augment with motion information to better estimate moving
foreground subjects in videos.  This can achieve better alignment, but requires
the entire dataset to be available at runtime and performs expensive alignment
procedures.  By contrast, our method learns an easier-to-store set of network
parameters, and can be applied to images in real-time.

More broadly, stereo depth estimation has been extensively
investigated.  Scharstein \etal \cite{Scharstein02ataxonomy} provide a survey
and evaluation of many methods for 2-frame stereo correspondence,
organized by matching, aggregation and optimization techniques.  In a creative
application of multiview stereo, Snavely \etal \cite{Snavely06} match across
views of many uncalibrated consumer photographs of the same scene to create
accurate 3D reconstructions of common landmarks. 

Machine learning techniques have also been applied in the stereo case, often
obtaining better results while relaxing the need for careful camera alignment
\cite{Konda13,Memisevic11,Yamaguchi12,Sinz04}.  Most relevant to this work
is Konda \etal \cite{Konda13}, who train a factored autoencoder on image
patches to predict depth from stereo sequences; however, this relies on the
local displacements provided by stereo.

There are also several hardware-based solutions for single-image depth
estimation.  Levin \etal \cite{Levin07} perform depth from defocus using a
modified camera aperture, while the Kinect and Kinect v2 
use active stereo and time-of-flight to capture depth.  Our method makes
indirect use of such sensors to provide ground truth depth targets
during training;
however, at test time our system is purely software-based, predicting depth from
RGB images.

\vspace{-3mm}
\section{Approach}
\vspace{-2mm}

\begin{figure}
\centering
\includegraphics[width=\linewidth]{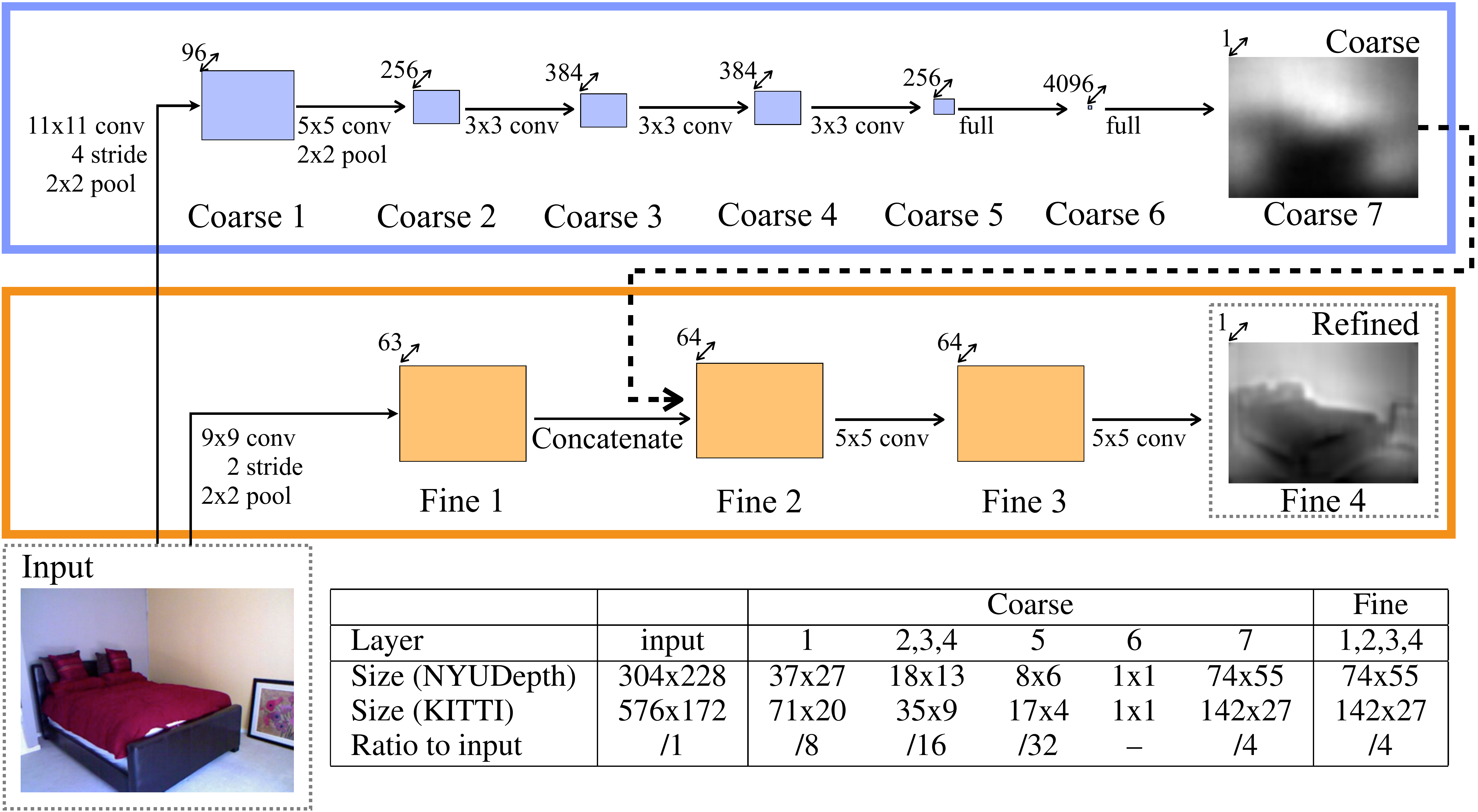}
\vspace{-8mm}
\caption{Model architecture.}
\vspace{-6mm}
\label{fig:arch}
\end{figure}

\vspace{-1mm}
\subsection{Model Architecture}
\vspace{-2mm}
Our network is made of two component stacks, shown in \fig{arch}.  A
coarse-scale network first predicts the depth of the scene at a global level.
This is then refined within local regions by a fine-scale network.  Both stacks
are applied to the original input, but in addition, the coarse network's output is
passed to the fine network as additional first-layer image features.  In this
way, the local network can edit the global prediction to incorporate
finer-scale details.

\vspace{-2mm}
\subsubsection{Global Coarse-Scale Network}
\label{sec:globalnet}
\vspace{-2mm}
The task of the coarse-scale network is to predict the overall depth map
structure using a global view of the scene.  The upper layers of this network
are fully connected, and thus contain the entire image in their field of view.
Similarly, the lower and middle layers are designed to combine information from
different parts of the image through max-pooling operations to a small spatial
dimension.  In so doing, the network is able to integrate a global
understanding of the full scene to predict the depth.  Such
an understanding is needed in the single-image case to make effective use of
cues such as vanishing points, object locations, and room alignment.  A local
view (as is commonly used for stereo matching) is insufficient to notice
important features such as these.

As illustrated in \fig{arch}, the
global, coarse-scale network contains five feature extraction layers of
convolution and max-pooling, followed by two fully connected layers.  The input,
feature map and output sizes are also given in \fig{arch}.  The final output
is at $1/4$-resolution compared to the input (which is itself downsampled
from the original dataset by a factor of 2), and corresponds to a center crop
containing most of the input (as we describe later, we lose a small
border area due to the first layer of the fine-scale network and image transformations).

\begin{figure}[h]
\centering
\vspace{-5mm}
\begin{tabular}{cc}
\includegraphics[scale=0.35]{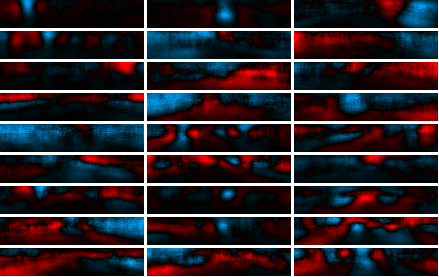}
&
\includegraphics[scale=0.35]{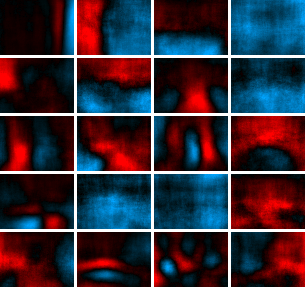}
\\
(a) & (b)
\end{tabular}
\vspace{-3mm}
\caption{Weight vectors from layer Coarse 7 (coarse output), for \enum{a} KITTI and \enum{b} NYUDepth.
Red is positive (farther) and blue is negative (closer); black is zero.
Weights are selected uniformly and shown in descending order by $l_2$ norm.  KITTI weights often show
changes in depth on either side of the road. NYUDepth weights often show wall positions and
doorways.}
\label{fig:weights}
\vspace{-3mm}
\end{figure}

Note that the spatial dimension of the output is larger than that of the
topmost convolutional feature map.  Rather than limiting the output to the
feature map size and relying on hardcoded upsampling before passing the
prediction to the fine network, we allow the top full layer to learn templates
over the larger area (74x55 for NYU Depth).  These are expected to be blurry, but will be 
better than the upsampled output of a 8x6 prediction (the top
feature map size); essentially, we allow the network to learn its own
upsampling based on the features.  Sample output weights
are shown in \fig{weights}

All hidden layers use rectified linear units for activations, with the
exception of the coarse output layer 7, which is linear.  Dropout is applied to the
fully-connected hidden layer 6.
The convolutional layers (1-5) of the coarse-scale network are pretrained on the
ImageNet classification task \cite{Deng09imagenet} --- while developing the model, we
found pretraining on ImageNet worked better than initializing randomly,
although the difference was not very large\footnote{When pretraining, we
stack two fully connected layers with 4096 - 4096 - 1000 output units each,
with dropout applied to the two hidden layers, as in \cite{Kriz12}.  We train the network using
random 224x224 crops from the center 256x256 region of each training image,
rescaled so the shortest side has length 256.  This model achieves a top-5
error rate of 18.1\% on the ILSVRC2012 validation set, voting with 2 flips and
5 translations per image.}.  

\vspace{-2mm}
\subsubsection{Local Fine-Scale Network}
\label{sec:localnet}
\vspace{-2mm}

After taking a global perspective to predict the coarse depth map, we make
local refinements using a second, fine-scale network.  The task of this
component is to edit the coarse prediction it receives to align with local
details such as object and wall edges.  The fine-scale network stack consists
of convolutional layers only, along with one pooling stage for the first layer
edge features.

While the coarse network sees the entire
scene, the field of view of an output unit in the fine network is 45x45 pixels
of input.  The convolutional layers are applied across feature maps at the
target output size, allowing a relatively high-resolution output at $1/4$ the
input scale.

More concretely, the coarse output is fed in as an additional low-level feature
map.  By design, the coarse prediction is the same spatial size as the output of
the first fine-scale layer (after pooling), and we concatenate the two
together (Fine 2 in \fig{arch}).  Subsequent layers maintain this size using
zero-padded convolutions.

All hidden units use rectified linear activations.  The
last convolutional layer is linear, as it predicts the target depth.  We train
the coarse network first against the ground-truth targets, then train the fine-scale
network keeping the coarse-scale output fixed (\ie when training the fine
network, we do not backpropagate through the coarse one).

\vspace{-2mm}
\subsection{Scale-Invariant Error}
\label{sec:sierror}
\vspace{-2mm}

The global scale of a scene is a fundamental ambiguity in depth prediction.  
Indeed, much of the error accrued using current elementwise metrics may
be explained simply by how well the mean depth is predicted.  For example,
Make3D trained on NYUDepth obtains 0.41 error using RMSE in log space (see
\tab{nyudresults}). However, using an oracle to substitute the mean log depth of
each prediction with the mean from the corresponding ground truth reduces the
error to 0.33, a 20\% relative improvement.  Likewise, for our system, these
error rates are 0.28 and 0.22, respectively.  Thus, just finding the average
scale of the scene accounts for a large fraction of the total error.

Motivated by this, we use a scale-invariant error to
measure the relationships between points in the scene, irrespective of the
absolute global scale.
For a predicted depth map $y$ and ground truth $y^*$, each with $n$ pixels
indexed by $i$,
we define the \emph{scale-invariant mean squared error} (in log space) as

\vspace{-6mm}
\begin{eqnarray}
\label{eqn:sierror}
D(y, y^*) & = & \frac{1}{n} \sum_{i=1}^n (\log y_i - \log y_i^* + \alpha(y,y^*))^2,
\end{eqnarray}
\vspace{-5mm}

where $\alpha(y,y^*) = \frac{1}{n} \sum_i(\log y_i^* - \log y_i)$ is the
value of $\alpha$ that minimizes the error for a given $(y, y^*)$.
For any prediction $y$, $e^\alpha$ is the scale that best aligns it to the ground
truth. All scalar multiples of $y$ have the same error, hence the scale invariance.

Two additional ways to view this metric are provided by the following equivalent
forms. Setting $d_i = \log y_i - \log y_i^*$ to be the difference between the
prediction and ground truth at pixel $i$, we have
\vspace{-1mm}
\begin{eqnarray}
\label{eqn:pairwise}
D(y, y^*) & = & \frac{1}{n^2} \sum_{i,j} \left( (\log y_i - \log y_j) - (\log y_i^* - \log y_j^*) \right)^2 
\end{eqnarray}
\vspace{-6mm}
\begin{eqnarray}
\label{eqn:pairdiff}
\hspace{26.7mm} & = & \frac1n \sum_i d_i^2 - \frac1{n^2} \sum_{i,j} d_i d_j ~~~ = ~~~ \frac1n \sum_i d_i^2 - \frac1{n^2} \left(\sum_i d_i \right)^2
\end{eqnarray}
\vspace{-5mm}

\eqn{pairwise} expresses the error by comparing relationships
\emph{between pairs} of pixels $i,j$ in the output: to have low error, each
pair of pixels in the prediction must differ in depth by an amount similar to that of the corresponding
pair in the ground truth.
\eqn{pairdiff} relates the metric to the original $l_2$ error,
but with an additional term, $-\frac{1}{n^2}\sum_{ij}d_id_j$, that credits
mistakes if they are in the same direction and penalizes them if they
oppose.  Thus, an imperfect prediction will have lower error when its mistakes
are consistent with one another.  The last part of \eqn{pairdiff} rewrites this
as a linear-time computation.

In addition to the scale-invariant error, we also measure the performance of
our method according to several error metrics have been proposed in prior works,
as described in \secc{experiments}.

\vspace{-2mm}
\subsection{Training Loss}
\vspace{-2mm}

In addition to performance evaluation, we also tried using the scale-invariant error as a training
loss.  Inspired by \eqn{pairdiff}, we set the per-sample training loss to
\vspace{-3mm}
\begin{eqnarray}
\label{eqn:loss}
L(y, y^*) & = & \frac1n \sum_i d_i^2 - \frac\lambda{n^2} \left(\sum_i d_i \right)^2
\end{eqnarray}
where $d_i = \log y_i - \log y_i^*$ 
and $\lambda \in [0,1]$.  Note the output of the network is $\log y$; that is,
the final linear layer predicts the log depth.
Setting $\lambda=0$ reduces to elementwise $l_2$,
while $\lambda=1$ is the scale-invariant error exactly.  We use
the average of these, \ie $\lambda=0.5$, finding that this produces 
good absolute-scale predictions while slightly improving qualitative output.

During training, most of the target depth maps will have some missing
values, particularly near object boundaries, windows and specular surfaces.  We
deal with these simply by masking them out and evaluating the loss only on
valid points, \ie we replace $n$ in \eqn{loss} with the number of pixels that
have a target depth, and perform the sums excluding pixels $i$ that have no
depth value.

\vspace{-3mm}
\subsection{Data Augmentation}
\vspace{-3mm}

We augment the training data with random online transformations
(values shown for NYUDepth) \footnote{For
KITTI, $s \in [1, 1.2]$, and rotations are not performed (images are
horizontal from the camera mount).}:

\vspace{-2mm}
\begin{itemize}[noitemsep]
\item \emph{Scale}:  Input and target images are scaled by $s \in [1, 1.5]$, and the depths are divided by $s$.
\item \emph{Rotation}:  Input and target are rotated by $r \in [-5, 5]$ degrees.
\item \emph{Translation}:  Input and target are randomly cropped to the sizes indicated in \fig{arch}.
\item \emph{Color}:  Input values are multiplied globally by a random RGB value $c \in [0.8, 1.2]^3$.
\item \emph{Flips}:  Input and target are horizontally flipped with 0.5 probability.
\end{itemize}
\vspace{-2mm}

Note that image scaling and translation do not preserve the world-space
geometry of the scene.  This is easily corrected in the case of scaling by
dividing the depth values by the scale $s$ (making the image $s$ times larger
effectively moves the camera $s$ times closer).  Although translations are not
easily fixed (they effectively change the camera to be incompatible with the
depth values), we found that the extra data they provided benefited the
network even though the scenes they represent were slightly warped.  The other transforms,
flips and in-plane rotation, are geometry-preserving.
At test time, we use a single center crop at scale 1.0 with no rotation or color transforms.

\vspace{-4mm}
\section{Experiments}
\vspace{-4mm}
\label{sec:experiments}

We train our model on the raw versions both NYU Depth v2 \cite{Silberman12} and KITTI \cite{Geiger2013IJRR}.
The raw distributions contain many additional images
collected from the same scenes as in the more commonly used small
distributions, but with no preprocessing; in particular, points for which there
is no depth value are left unfilled.  However, our model's natural ability to
handle such gaps as well as its demand for large training sets make these
fitting sources of data.

\vspace{-3mm}
\subsection{NYU Depth}
\vspace{-2mm}

The NYU Depth dataset \cite{Silberman12} is composed of 464 indoor scenes,
taken as video sequences using a Microsoft Kinect camera.  We use the official
train/test split, using 249 scenes for training and 215 for testing, and
construct our training set using the raw data for these scenes.  RGB inputs are
downsampled by half, from 640x480 to 320x240.  Because the
depth and RGB cameras operate at different variable frame rates, we associate
each depth image with its closest RGB image in time, and throw away frames
where one RGB image is associated with more than one depth (such a one-to-many
mapping is not predictable).  We use the camera projections provided with the
dataset to align RGB and depth pairs; pixels with no depth value are left
missing and are masked out. To remove many invalid regions caused by windows,
open doorways and specular surfaces we also mask out depths equal to the
minimum or maximum recorded for each image.

The training set has 120K unique images, which we shuffle into a list of
220K after evening the scene distribution (1200 per scene).
We test on the 694-image
NYU Depth v2 test set (with filled-in depth values).
We train coarse network for 2M samples using SGD with batches of size 32.
We then hold it fixed and train the fine network for 1.5M samples (given
outputs from the already-trained coarse one).  Learning rates are: 0.001 for
coarse convolutional layers 1-5, 0.1 for coarse full layers 6 and 7, 0.001 for
fine layers 1 and 3, and 0.01 for fine layer 2.  These ratios were found by
trial-and-error on a validation set (folded back into the training set for our
final evaluations), and the global scale of all the rates was tuned to a
factor of 5.  Momentum was 0.9.

\vspace{-3mm}
\subsection{KITTI}
\vspace{-3mm}

The KITTI dataset \cite{Geiger2013IJRR} is composed of several outdoor scenes
captured while driving with car-mounted cameras and depth sensor.  We use 56
scenes from the ``city,'' ``residential,'' and ``road'' categories of the raw
data.  These are split into 28 for training and 28 for testing.  The RGB images
are originally 1224x368, and downsampled by half to form
the network inputs.

The depth for this dataset is sampled at irregularly spaced points, captured at
different times using a rotating LIDAR scanner.  When constructing the ground
truth depths for training, there may be conflicting values; since the RGB
cameras shoot when the scanner points forward, we resolve conflicts at each
pixel by choosing the depth recorded closest to the RGB capture time.  Depth is
only provided within the bottom part of the RGB image, however we feed the
entire image into our model to provide additional context to the global
coarse-scale network (the fine network sees the bottom crop corresponding to
the target area).

The training set has 800 images per scene.  We exclude shots where the car is
stationary (acceleration below a threshold) to avoid duplicates.  Both left and
right RGB cameras are used, but are treated as unassociated shots.  The
training set has 20K unique images, which we shuffle into a list of 40K
(including duplicates) after evening the scene distribution.
We train the coarse model first for 1.5M samples, then the fine model for 1M.
Learning rates are the same as for NYU Depth.

\vspace{-3mm}
\subsection{Baselines and Comparisons}
\vspace{-3mm}
\label{sec:baselines}

We compare our method against Make3D trained on the same datasets, as well as
the published results of other current methods \cite{Ladicky14,Karsch14}.  As
an additional reference, we also compare to the mean depth image computed
across the training set.
We trained Make3D on KITTI using a subset of 700 images (25 per scene), as the
system was unable to scale beyond this size.  Depth targets were filled in
using the colorization routine in the NYUDepth development kit.  For NYUDepth, we used the common distribution
training set of 795 images.
We evaluate each method using several errors from prior
works, as well as our scale-invariant metric:

{
\small
\begin{tabular}{l | l}

Threshold:  \% of $y_{i}$ s.t. $\max(\frac{y_i}{y_i^*},\frac{y_i^*}{y_i}) = \delta < thr$
 & RMSE (linear):  $\sqrt{\frac1{|T|}\sum_{y\in T}||y_i - y_i^*||^2}$
\\
Abs Relative difference:  $\frac1{|T|}\sum_{y\in T}|y - y^*| / y^*$
 & RMSE (log):  $\sqrt{\frac1{|T|}\sum_{y\in T}||\log y_i - \log y_i^*||^2}$ 
\\
Squared Relative difference:  $\frac1{|T|}\sum_{y\in T}||y - y^*||^2 / y^*$
 & RMSE (log, scale-invariant):  The error \eqn{sierror}

\end{tabular}
}

Note that the predictions from Make3D and our network correspond to slightly
different center crops of the input.  We compare them on the intersection of
their regions, and upsample predictions to the full original input resolution
using nearest-neighbor.
Upsampling negligibly affects performance compared to downsampling
the ground truth and evaluating at the output resolution.
\footnote{On NYUDepth, log RMSE is 0.285 vs 0.286 for upsampling and downsampling, respectively, and
scale-invariant RMSE is 0.219 vs 0.221.
The intersection is 86\% of the network region and 100\% of Make3D
for NYUDepth, and 100\% of the network and 82\% of Make3D for KITTI.
}

\vspace{-4mm}
\section{Results}
\label{sec:results}
\vspace{-2mm}

\vspace{-2mm}
\subsection{NYU Depth}
\vspace{-3mm}

Results for NYU Depth dataset are provided in \tab{nyudresults}.  As explained
in \secc{baselines}, we compare against the data mean and Make3D as
baselines, as well as Karsch \etal \cite{Karsch14} and Ladicky \etal
\cite{Ladicky14}.
(Ladicky \etal uses a joint model which is trained using both depth and semantic labels).
Our system achieves the best performance on all metrics, obtaining an average
35\% relative gain compared to the runner-up.  Note that our system is trained
using the raw dataset, which contains many more example instances than
the data used by other approaches, and is able to effectively
leverage it to learn relevant features and their associations.

This dataset breaks many assumptions made by Make3D, particularly
horizontal alignment of the ground plane; as a result, Make3D has
relatively poor performance in this task.  Importantly, our method improves
over it on both scale-dependent and scale-invariant metrics, showing that our
system is able to predict better relations as well as better means.

\begin{figure}[h]
\vspace{-2mm}
\centering
\hspace{-2mm}\includegraphics[width=\linewidth]{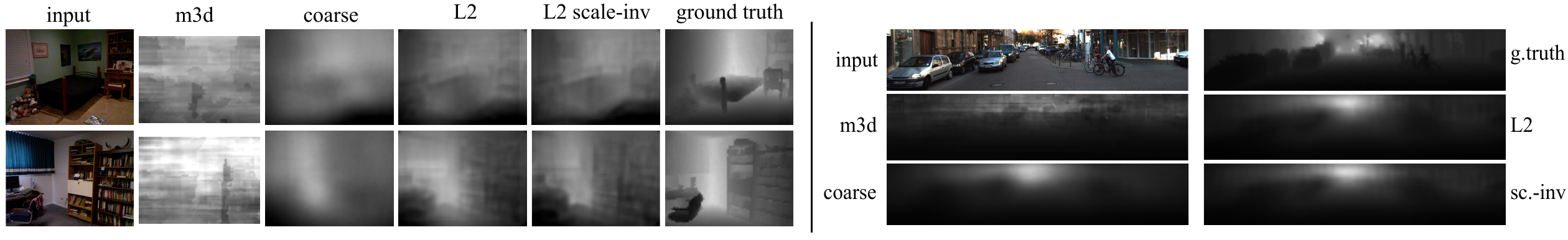}
\vspace{-4mm}
\caption{Qualitative comparison of Make3D, our method trained with $l_2$ loss ($\lambda=0$),
and our method trained with both $l_2$ and scale-invariant loss ($\lambda=0.5$).}
\label{fig:compare}
\vspace{-4mm}
\end{figure}

Qualitative results are shown on the left side of \fig{examples}, sorted
top-to-bottom by scale-invariant MSE.  Although the fine-scale network does
not improve in the error measurements, its effect is clearly
visible in the depth maps --- surface boundaries have sharper transitions,
aligning to local details.  However, some texture edges are sometimes also
included.  \fig{compare} 
compares Make3D as well as outputs from our network trained with losses
using $\lambda=0$ and $\lambda=0.5$.  While we did not
observe numeric gains using $\lambda=0.5$ over $\lambda=0$, it did produce
slight qualitative improvements in the more detailed outputs.

\vspace{-1mm}

\begin{table}[h]
\centering
\small
\begin{tabular}{|l||cccc|cc|c|}
\hline
                                 & Mean  & \hspace{-2mm} Make3D  &  \hspace{-2mm}Ladicky{\it\&al} & \hspace{-2mm}Karsch{\it\&al} & Coarse & \hspace{-1mm}Coarse + Fine & \\
\hline                           
\hline
threshold $\delta < 1.25$        &  0.418    &  0.447  &   0.542  &  --     & \bf{0.618}   &   0.611 & {\small higher}  \\  
threshold $\delta < 1.25^2$      &  0.711    &  0.745  &   0.829  &  --     & \bf{0.891}   &   0.887 & {\small is}  \\  
threshold $\delta < 1.25^3$      &  0.874    &  0.897  &   0.940  &  --     & 0.969   &   \bf{0.971} & {\small better} \\  
\hline                           
abs relative difference          &  0.408    &  0.349  &    --    &  0.350 &  0.228   &   \bf{0.215}  &  \\  
sqr relative difference          &  0.581    &  0.492  &    --    &  --    &  0.223   &   \bf{0.212} & {\small lower}   \\  
 RMSE (linear)                   &  1.244    &  1.214  &    --    &  1.2   &  \bf{0.871}   &   0.907  & {\small is}  \\  
 RMSE (log)                      &  0.430    &  0.409  &    --    &  --    &  \bf{0.283}   &   0.285   & {\small better} \\  
 RMSE (log, scale inv.)          &  0.304    &  0.325  &    --    &  --    &  0.221   &   \bf{0.219}   & \\  
\hline
\end{tabular}
\vspace{-3mm}
\caption{Comparison on the NYUDepth dataset}
\label{tab:nyudresults}
\end{table}

\vspace{-3mm}

\subsection{KITTI}
\vspace{-2mm}

We next examine results on the KITTI driving dataset.  Here, the Make3D
baseline is well-suited to the dataset, being composed of horizontally aligned
images, and achieves relatively good results.  Still, our method improves over
it on all metrics, by an average 31\% relative gain.  Just as importantly,
there is a 25\% gain in both the scale-dependent and
scale-invariant RMSE errors, showing there is substantial improvement in the
predicted structure.  Again, the fine-scale network does not improve much over
the coarse one in the error metrics, but 
differences between the two can be seen in the qualitative outputs.

\begin{figure}[t]
\centering
\hspace{-2mm}\includegraphics[width=1.1\linewidth]{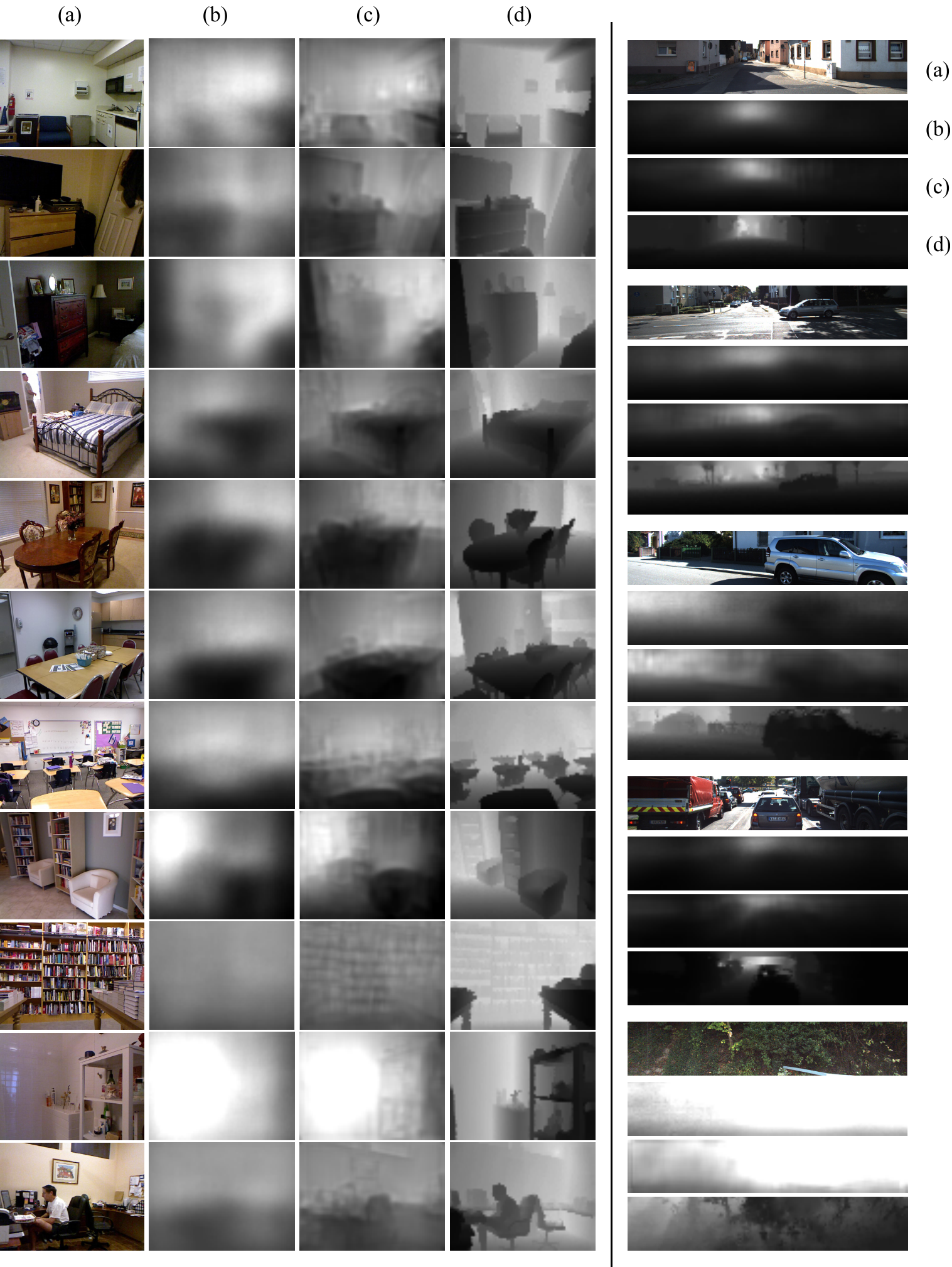}
\vspace{-4mm}
\caption{Example predictions from our algorithm.  NYUDepth on left, KITTI on right.  For each image, we show (a) input, (b) output of coarse network, (c) refined output of fine network, (d) ground truth.  The fine scale network edits the coarse-scale input to better align with details such as object boundaries and wall edges.  Examples are sorted from best (top) to worst (bottom).}
\label{fig:examples}
\end{figure}

The right side of \fig{examples} shows examples of predictions, again sorted
by error.  The fine-scale
network produces sharper transitions here as well,
particularly near the road edge.  However, the changes are somewhat limited.
This is likely caused by uncorrected alignment issues between the depth map and
input in the training data, due to the rotating scanner setup.  This
dissociates edges from their true position, causing the network to average
over their more random placements.  \fig{compare} shows Make3D performing much
better on this data, as expected, while using the scale-invariant error
as a loss seems to have little effect in this case.

\vspace{-1mm}

\begin{table}[h]
\centering
\small
\begin{tabular}{|l||cc|cc|c|}
\hline
                                 & Mean  &  Make3D  & Coarse & Coarse + Fine  & \\
\hline                           
\hline                           
threshold $\delta < 1.25$        &   0.556   &     0.601   &  0.679     &    \bf{0.692}  &  {\small higher} \\  
threshold $\delta < 1.25^2$      &   0.752   &     0.820   &  0.897     &    \bf{0.899}  & {\small is} \\  
threshold $\delta < 1.25^3$      &   0.870   &     0.926   &  \bf{0.967}     &    \bf{0.967}   & {\small better} \\  
\hline                           
abs relative difference          &   0.412     &   0.280   &  0.194     &    \bf{0.190}   &  \\  
sqr relative difference          &   5.712     &   3.012   &  1.531     &    \bf{1.515}     & {\small lower} \\  
 RMSE (linear)                   &   9.635     &   8.734   &  7.216     &    \bf{7.156}     & {\small is} \\  
 RMSE (log)                      &   0.444     &   0.361   &  0.273     &    \bf{0.270}  &  {\small better} \\  
 RMSE (log, scale inv.)          &   0.359     &   0.327   &  0.248     &    \bf{0.246}  &   \\  
\hline
\end{tabular}
\vspace{-2mm}
\caption{Comparison on the KITTI dataset.}
\label{tab:kittiresults}
\end{table}

\vspace{-2mm}

\vspace{-5mm}
\section{Discussion}
\vspace{-3mm}

Predicting depth estimates from a single image is a challenging task.  Yet by
combining information from both global and local views, it can be performed
reasonably well.  Our system accomplishes this through the use of two deep
networks, one that estimates the global depth structure, and another that
refines it locally at finer resolution.  We achieve a new state-of-the-art on
this task for NYU Depth and KITTI datasets, having effectively leveraged the
full raw data distributions.

In future work, we plan to extend our method to incorporate further 3D geometry
information, such as surface normals.  Promising results in normal map
prediction have been made by Fouhey \etal \cite{fouhey2013data}, and
integrating them along with depth maps stands to improve overall performance
\cite{Saxena08}.  We also hope to extend the depth maps to the full original
input resolution by repeated application of successively finer-scaled local
networks.

\clearpage

{
\bibliographystyle{ieee}
\bibliography{depth}
}

\end{document}